\documentclass[conference]{IEEEtran}
\IEEEoverridecommandlockouts
\usepackage{cite}
\usepackage{amsmath,amssymb,amsfonts}
\usepackage{algorithm}
\usepackage{algorithmic}
\usepackage{graphicx}
\usepackage{textcomp}
\usepackage{xcolor}
\usepackage{verbatim} 
\usepackage{makecell} 
\usepackage{diagbox} 
\usepackage{multirow}
\usepackage{CJKutf8}
\usepackage{booktabs}

\def\BibTeX{{\rm B\kern-.05em{\sc i\kern-.025em b}\kern-.08em
    T\kern-.1667em\lower.7ex\hbox{E}\kern-.125emX}}

\def\methodname{AMTSum}
\def\clsdatasetname{AspectSent}
\begin{document}

\title{Aspect-based Meeting Transcript Summarization: A Two-Stage Approach with Weak Supervision on Sentence Classification}

\author{\IEEEauthorblockN{Zhongfen Deng}
\IEEEauthorblockA{\textit{Department of Computer Science}\\
\textit{University of Illinois Chicago}\\
Chicago, Illinois 60607\\
Email: zdeng21@uic.edu}
\and
\IEEEauthorblockN{Seunghyun Yoon\\ Trung Bui\\ Franck Dernoncourt\\ Quan Hung Tran}
\IEEEauthorblockA{\textit{Adobe Research}\\
\textit{}
Email: syoon@adobe.com\\
bui@adobe.com\\
dernonco@adobe.com\\
qtran@adobe.com}
\and
\IEEEauthorblockN{Shuaiqi Liu}
\IEEEauthorblockA{\textit{Department of Computing}\\
\textit{The Hong Kong Polytechnic University}\\
HongKong, China\\
Email: cssqliu@comp.polyu.edu.hk}
\and
\IEEEauthorblockN{Wenting Zhao\\ Tao Zhang\\ Yibo Wang}
\IEEEauthorblockA{\textit{Department of Computer Science}, \\
\textit{University of Illinois Chicago}\\
Chicago, Illinois 60607\\
Email: wzhao41@uic.edu,
tzhang90@uic.edu,
ywang633@uic.edu}
\and
\IEEEauthorblockN{Philip S. Yu}
\IEEEauthorblockA{\textit{Department of Computer Science}\\
\textit{University of Illinois Chicago}\\
Chicago, Illinois 60607\\
Email: psyu@uic.edu}
}

\maketitle

\begin{abstract} 
Aspect-based meeting transcript summarization aims to produce multiple summaries, each focusing on one aspect of content in a meeting transcript.
It is challenging as sentences related to different aspects can mingle together, and those relevant to a specific aspect can be scattered throughout the long transcript of a meeting. The traditional summarization methods produce one summary mixing information of all aspects, which cannot deal with the above challenges of aspect-based meeting transcript summarization. In this paper, we propose a two-stage method for aspect-based meeting transcript summarization. To select the input content related to specific aspects, we train a sentence classifier on a dataset constructed from the AMI corpus with pseudo-labeling. Then we merge the sentences selected for a specific aspect as the input for the summarizer to produce the aspect-based summary. Experimental results on the AMI corpus outperform many strong baselines, which verifies the effectiveness of our proposed method.

\end{abstract}


\section{Introduction}
With the increase of online video meetings, the need for meeting summarization is emerging. The meeting summary usually needs to summarize the discussion content in multiple aspects. 
Traditional summarization task usually produces a single overall summary for an input document such as news, scientific article, customer review, etc. It cannot meet the requirement of summarizing meeting content in specific aspects (e.g., problems, decisions). Therefore, we propose the aspect-based meeting transcript summarization, which aims to generate informative, fluent, non-redundant summaries for different aspects respectively. 
To achieve this goal, we need to solve two challenging issues: 1) Meeting transcript sentences related to different aspects can mingle together, which makes it difficult to generate a summary for a specific aspect. 2) The sentences related to an aspect can be scattered throughout the meeting 
which includes thousands of words.

The current models for traditional summarization task cannot cope with these challenges in aspect-based meeting transcript summarization.
Therefore, we propose a two-stage method named {\methodname} for Aspect-based Meeting Transcript Summarization to address these challenges. It first selects the input sentences related to each aspect, and then merges the selected sentences as the input of the summarizer to produce the aspect-based summary. 
Due to the unavailability of aspect labels for each sentence in the meeting transcript, we first construct a pseudo-labeled sentence classification dataset called {\clsdatasetname} from the AMI corpus \cite{carletta2006ami} by utilizing the state-of-the-art sentence embedding models. In the first stage, we design and train a multi-label classifier on {\clsdatasetname} to identify sentences related to each aspect. 
In the second stage, the sentences selected for the same aspect are merged with a special token as the input of the summarizer to produce an aspect-based summary.  
We train one abstractive summarization model to generate summaries for all aspects. 
We train and evaluate our method and various extractive and abstractive summarization models on the AMI corpus. Experimental results show that our method outperforms the competitive baselines, including large pre-trained language models, which verifies its effectiveness. 

To sum up, our main contributions are as follows.
\begin{itemize}
\item We are the first to propose the task of aspect-based meeting transcript summarization, which aims at producing the summary focusing on each aspect of the meeting content individually.
\item We propose a two-stage method for aspect-based meeting transcript summarization, which includes a weakly supervised multi-label sentence classifier in the first stage and a summarizer in the second stage. 

\item We construct a pseudo-labeled dataset from the AMI corpus to train the sentence classifier.
\item Experimental results on the AMI corpus validate the effectiveness of our proposed method.
\end{itemize}

\section{Related Work}
\subsection{Meeting Summarization}
Traditional summarization models focus on written documents such as news, articles, product reviews, etc. With the increase of online meetings in recent years, summarization for multi-party conversations or meetings receives a lot of attention from the research community. As different types of meetings or conversations have their characteristics, there does not exist a universal model which can be applied to different domains or types of meetings. \cite{krishna-etal-2021-generating} propose a model to produce a summary of a medical conversation between a doctor and a patient. \cite{liu2019automatic} design a model for summarizing customer service calls by using the key points in the dialog. 
\cite{zhu-etal-2020-hierarchical} propose a hierarchical network including a word-level and turn-level transformer to tackle the lengthy input meeting transcript to generate a summary. 
There are also research works using auxiliary information to help generate a summary for a meeting, such as domain terminology \cite{koay-etal-2020-domain}, discourse structure and relations \cite{feng2020dialogue}. \cite{zhang-etal-2021-exploratory-study} investigate three strategies to deal with the long meeting transcript and find that the retrieve-then-summarize method works best for meeting summarization. Another strategy is the sliding window method proposed by \cite{koay-etal-2021-sliding}. \cite{zhong2022dialoglm} design a pre-training model for long dialogue summarization.
\cite{zhuang-etal-2021-weakly,liu-etal-2021-coreference,liu-chen-2022-entity} utilize different technique (e.g., attention) or auxiliary information (e.g., coreference, entity) to improve dialog summarization performance. \cite{manuvinakurike-etal-2021-incremental} develop an incremental temporal summarization dataset for multi-party meetings.
All these models produce an overall summary of a dialogue or meeting. Conversely, our task and method aim to summarize each aspect of the meeting content individually.

\subsection{Aspect-based Summarization}
Aspect-based summarization aims to produce a summary of an input document for an abstract aspect. \cite{krishna-srinivasan-2018-generating,frermann-klementiev-2019-inducing} design aspect-based summarization models for news articles, which use the categories of news as the aspects. Aspect-based summarization has also been explored in the product review domain. \cite{angelidis-lapata-2018-summarizing} propose an extractive aspect-based opinion summarization model to form opinion summaries from multiple product reviews.
Recently, \cite{tan-etal-2020-summarizing} introduce external knowledge such as ConceptNet and Wikipedia into the aspect-based summarization model for the synthetic data MA-News, which is constructed from the CNN/DailyMail dataset.  

\subsection{Query-based Summarization}
Summarization models producing a summary of a document (e.g., news, wikipedia article, debate) for a given query (i.e., a natural language question) are also studied by many researchers \cite{daume2006bayesian,wang2013sentence,nema-etal-2017-diversity,kulkarni2020aquamuse,ishigaki2020neural,laskar2020query,zhu2022transforming}. Recently, \cite{zhong-etal-2021-qmsum} define a new task of query-based multi-domain meeting summarization which aims to produce a single piece of text that answers a specific query.
They propose a benchmark QMSum for this task by annotating applicable input queries for each meeting. Specifically, QMSum uses a query schema list to guide the annotators to generate different queries for different meetings, which is labor-intensive.
Different from QMSum, our work aims at generating 
multiple aspect-based summaries covering the main content of the meeting with complete and multi-perspective information, it does not require input queries.

All the above models are either generating one overall summary of a meeting or generating a summary of a written document for a specific query or aspect. 
On the contrary, our task of aspect-based meeting transcript summarization aims to summarize each aspect of the meeting conversation individually.
To our best knowledge, there are no previous works for this task.

\section{Problem Formulation}
Given the transcript of a meeting and several aspects of this meeting, such as problems, actions, and decisions, the goal is to generate the summary individually for each aspect of the meeting content.
Assume the meeting transcript is denoted as $T = (w_1, w_2,..., w_L)$, $L$ is the length of the meeting transcript, the aspects for the meeting are denoted as $A = (a_1, a_2,..., a_m)$, $m$ is the number of aspects which exist in the meeting. The model aims to generate summaries for the corresponding aspects denoted as $S = (S_1, S_2, ..., S_m)$, each of the summaries can have a different length.

\section{Methodology}\label{sec:method}
\begin{figure*}[t]
    \centering
    \includegraphics[scale=0.65]{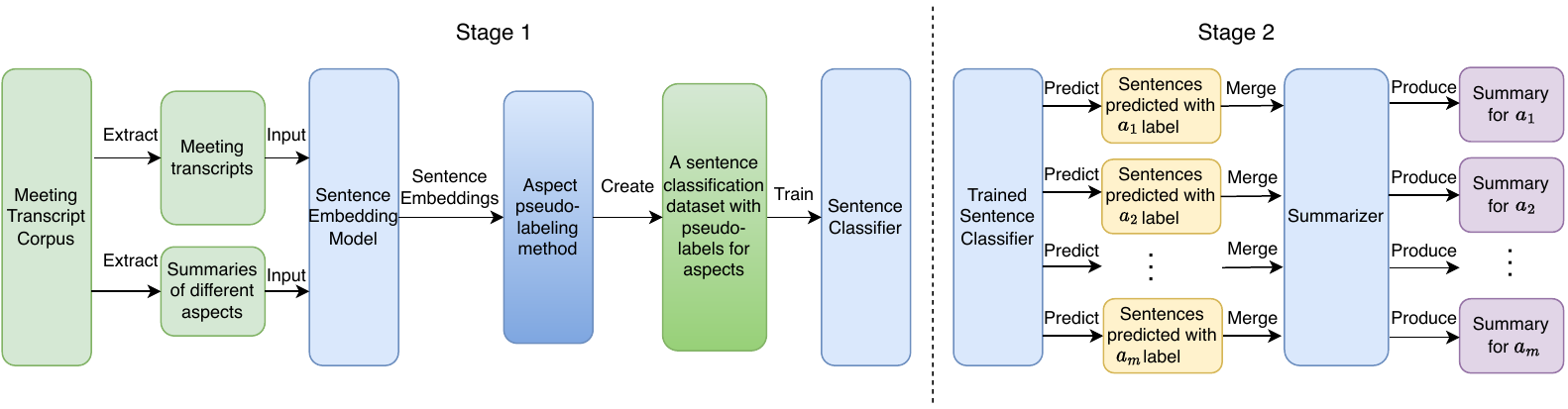}\vspace{-0.1in}
    \caption{The overview of our proposed method for aspect-based meeting transcript summarization.}\vspace{-0.25in}
    \label{fig:method_overview}
\end{figure*}

The content of different aspects in a meeting transcript mingle with each other, and the sentences related to a specific aspect can be scattered in the long meeting transcript, which makes it difficult to produce summaries for different aspects. Therefore, we design a two-stage method called {\methodname} to produce aspect-based summaries for a meeting transcript. Fig. \ref{fig:method_overview} shows the overview of {\methodname}.

\subsection{{\methodname}-Stage 1}
In stage 1, our approach involves several steps. Firstly, we extract meeting transcripts along with their corresponding summaries for different aspects from the AMI corpus. Next, we design a pseudo-labeling method to create aspect labels for each sentence in the meeting transcripts with the help of a sentence embedding model. It is utilized to generate embeddings for the sentences in both the meeting transcripts and aspect-specific summaries. Lastly, we construct a dataset with these aspect labels and train a sentence classifier, which predicts the relevance of a sentence to a specific aspect.

\begin{figure}[t]
    \centering
    \includegraphics[scale=0.70]{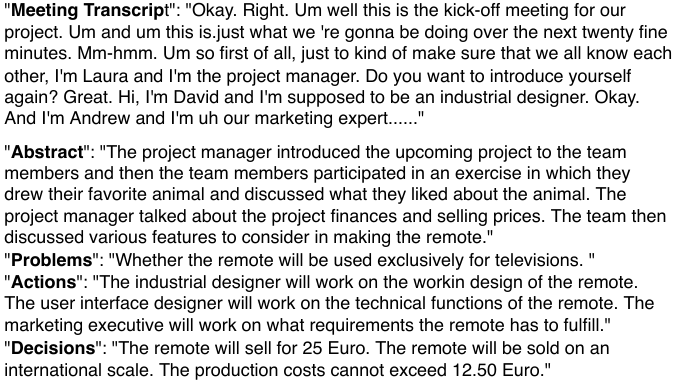}\vspace{-0.1in}
    \caption{One example of the extracted meeting transcripts and aspect-based summaries.}\vspace{-0.25in}
    \label{fig:example_meetingtrans}
\end{figure}

\noindent{\bf $\bullet$ \emph{Extracting Meeting Transcripts and Aspect-based Summaries}}
The most commonly used dataset for meeting transcript summarization is the AMI corpus \cite{carletta2006ami}, which contains many different forms of annotation for the content of the meeting, such as named entity annotation. We aim to utilize its transcripts for aspect-based meeting summarization. Therefore, we only extract the transcripts and manually annotated summaries for abstract, problems, actions, and decisions from the original AMI corpus. For each meeting, the extracted transcript and summaries are saved as a dictionary in a JSON file. Fig. \ref{fig:example_meetingtrans} shows one example of such files.

\noindent{\bf $\bullet$ \emph{Pseudo-labeling Method for Aspects}}
To create pseudo aspect labels for each sentence in a meeting transcript for training the sentence classifier (i.e., weakly-supervised learning for sentence classifier), we first utilize the recent state-of-the-art sentence embedding model SimCSE-BERT\textsubscript{base} and SimCSE-RoBERTa\textsubscript{large} \cite{gao-etal-2021-simcse} to learn representation for each sentence in the meeting transcript and the sentence in the summaries of different aspects. Then we design an algorithm to label each sentence by using the learned sentence embeddings. 
Each sentence will have $m$ labels, which correspond to the $m$ aspects in the meeting. The details of labeling aspects for each sentence are shown in Algorithm \ref{alg:algorithm}. Besides the similarity score, we also consider the length of each sentence in the meeting transcript and the length of reference summary for different aspect as shown in line 6 of Algorithm \ref{alg:algorithm} when labeling aspects. Because very short sentences are usually irrelevant to any aspect.

\begin{algorithm}[tb]
\small
\caption{Aspect labeling method for one meeting transcript}
\label{alg:algorithm}
\textbf{Input}: Embeddings for sentences in the meeting transcript and for aspect-based summaries, the threshold $\alpha$\\
\textbf{Output}: Sentences with aspect labels
\begin{algorithmic}[1] 
\STATE Let $SentsWithLabels=[ ]$.
\FOR{each sent $S$ in $Sents$}
\STATE Set all aspect labels $S_{a1}, S_{a2},...,S_{am}$ as zeros.
\FOR{aspect $ai$ in all aspects}
\STATE Calculate semantic similarity $Sim_i$ between the embedding of $S$ and $ai$.
\IF {$Sim_i > \alpha$ and length of $S>4$ and length of $ai>6$}
\STATE Set $S_{ai}=1$.
\ENDIF
\ENDFOR
\STATE Add $S$ into $SentsWithLabels$.
\ENDFOR
\STATE \textbf{return} $SentsWithLabels$
\end{algorithmic}
\end{algorithm}

\noindent{\bf $\bullet$ \emph{Dataset Construction and Sentence Classification Model}}\label{sec:sentclsdataset_construction}
After obtaining the aspect labels for sentences in all meeting transcripts, we use them to construct a sentence classification dataset called \textbf{{\clsdatasetname}} which contains more than 80,000 sentence examples. 
We design a multi-label classifier on top of BERT\textsubscript{base} \cite{devlin2018bert} to identify the sentences related to different aspects. 
Each sentence in the meeting transcript is provided as input to this model, which aims to predict aspect labels for each sentence,
i.e., whether the current sentence is relevant to a specific aspect or several aspects. Specifically, the classifier uses BERT\textsubscript{base} as the backbone, added with a dropout layer and a linear layer.
Sigmoid activation is utilized to produce the probability of relevance of each sentence with regard to the aspects. Binary Cross Entropy serves as the loss function during classifier training.  
Although trained on the constructed dataset with pseudo aspect labels, the classifier helps select the most relevant sentences for the summarizer to produce better aspect-based summaries. The effectiveness of the sentence classifier will be verified in the experimental results section.

\subsection{{\methodname}-Stage 2}
During the second stage, the trained sentence classifier is utilized to select the relevant sentences for each of the aspects in the meeting transcript. Those sentences are merged with a special token for each aspect as the input for the summarizer, 
which finally produces a summary for each aspect. 

\noindent{\bf $\bullet$ \emph{Sentence Selection}}
We use the trained sentence classifier to predict the aspect labels for each sentence in the meeting transcript.  
Sentences sharing the same predicted aspect label are grouped together and merged while preserving their original order.
This process generates $m$ filtered transcripts, each corresponding to a specific aspect. Additionally, a special token representing each aspect is added to both the filtered transcript and the target summary. This helps the summarizer distinguish which aspect-based summary it needs to produce for the given input. 
Moreover, this approach
helps augment the dataset size $m-1$ times which will be helpful for training the summarizer, especially in the low-resource settings (i.e., the limited number of annotated summarization training examples).

\noindent{\bf $\bullet$ \emph{Aspect-based Summaries Generation}}
Pre-trained language models have exhibited their strength in various natural language processing tasks in recent years. For the summarization task, more and more powerful pre-trained sequence-to-sequence models are emerging and designed to help improve the performance of generation and summarization, such as Pegasus\cite{zhang2020pegasus}, BART\cite{lewis-etal-2020-bart}, T5\cite{raffel2020exploring}, LED\cite{beltagy2020longformer} and so on. Any of these models can be applied as the summarizer in our method. We adopt the BART\textsubscript{large} model as the summarizer here because of its better performance in aspect-based meeting transcript summarization. The cross-entropy loss of the predicted tokens with respect to the ground-truth tokens in the reference summary is used to train the summarizer. We train a single summarizer to produce summaries for different aspects, as shown in stage 2 of Fig. \ref{fig:method_overview}.

\section{Experiments}
\subsection{Datasets}
\noindent{\bf $\bullet$ \emph{Dataset for Aspect-based Meeting Transcript Summarization}}
We run experiments on a publicly available dataset called AMI corpus \cite{carletta2006ami}. It is a dataset containing information for many different NLP tasks. To make it applicable to our task, we did some preprocessing to obtain the information needed for training aspect-based summarization model. Specifically, we first extract all the sentences in each meeting and also extract four aspect-based summaries for this meeting and store them as a JSON file for the next step. The number of meetings in this dataset is limited. We show some statistics of the processed dataset for our task in Table \ref{table:dataset_statistics}. To further evaluate our method, we construct another test set called ICSI-Test from the ICSI corpus \cite{janin2003icsi} by extracting the transcripts and corresponding "Problem" and "Decision" summaries. This constructed ICSI-Test set contains 61 testing examples.
\begin{table}[h]
\small
\centering
\resizebox{.48\textwidth}{!}{
\begin{tabular}{p{1.80cm}<{\centering} p{1.80cm}<{\centering} p{1.80cm}<{\centering} p{1.80cm}<{\centering}} 
 \hline
 \textbf{\#AMI-Train} & \textbf{\#AMI-Val} & \textbf{\#AMI-Test} & \textbf{\#ICSI-Test} \\ [0.5ex] 
 \hline
 100 & 21 & 21 & 61 \\ [0.5ex]
\hline
\end{tabular}
}
\caption{Statistics of processed AMI meetings and constructed ICSI-Test set for aspect-based summarization.}
\label{table:dataset_statistics}
\end{table}

\noindent{\bf $\bullet$ \emph{Dataset for Sentence Classification}}
As stated in Section \ref{sec:sentclsdataset_construction}, we construct a sentence classification dataset from the AMI corpus by creating pseudo aspect labels for each sentence in the meeting transcript. We keep the same split of training, validation, and test set as in the above processed AMI corpus for summarization. In other words, for the sentence classification dataset \textbf{{\clsdatasetname}}, we create sentences with aspect labels for the training, validation, and test set from the meeting transcripts in the corresponding set of the processed AMI corpus. The statistics of \textbf{{\clsdatasetname}} are shown in Table \ref{table:sentclsdataset_statistics}.
We can see that most of the sentences in the meeting transcripts are irrelevant to any of the aspects in the meeting, and the numbers of sentences with different aspect labels also vary. 
\begin{table}[h]
\small
\centering
\resizebox{.45\textwidth}{!}{%
\begin{tabular}{l |p{0.50cm}<{\centering} p{0.85cm}<{\centering} p{0.85cm}<{\centering} p{0.70cm}<{\centering} p{0.85cm}<{\centering} p{0.80cm}<{\centering}} 
 \hline
 \diagbox{}{\#S} & \textbf{Total} & \textbf{Abstract} & \textbf{Problem} & \textbf{Action} & \textbf{Decision} & \textbf{Irrelevant} \\ [0.5ex] 
 \hline
 Train & 56,408 & 942 & 1,605 & 419 & 2,225 & 51,217 \\ [0.5ex]
 Val & 11,703 & 171 & 165 & 77 & 378 & 10,912 \\ [0.5ex]
 Test & 13,761 & 182 & 421 & 81 & 407 & 12,670 \\ [0.5ex]
\hline
 Total & 81,872 & 1,295 & 2,191 & 577 & 3,010 & 74,799 \\ [0.5ex]
\hline
\end{tabular}
}
\caption{The statistics of the constructed sentence classification dataset \textbf{{\clsdatasetname}}. \#S means the number of sentence examples.}
\label{table:sentclsdataset_statistics}
\end{table}

\subsection{Evaluation Metrics}
We adopt the ROUGE F\textsubscript{1} scores \cite{lin-2004-rouge}, which include the overlap of unigrams (R-1), bigrams (R-2), and longest common subsequence (R-L)\footnote{github.com/falcondai/pyrouge/} to evaluate the performance of different summarization models.
Precision, Recall, and F1-score are used to evaluate the classifier's performance in predicting multiple aspect labels for each sentence in the meeting transcript.

\subsection{Baselines}
We compare our model with the state-of-the-art pre-trained language models, including T5, Pegasus, LED, and BART. They are trained in the preprocessed AMI dataset as described in the dataset section. Extractive models, including TextRank \cite{mihalcea2004textrank} and LexRank \cite{erkan2004lexrank} are also compared. 
Since ChatGPT\footnote{https://chat.openai.com/} 
exhibits great power in natural language generation, we also compare with it by designing appropriate prompts (refer to Table \ref{table:case_study_instructions}) to generate aspect-based summaries for meetings.

\subsection{Experimental Setting}
We use BART\textsubscript{large} as the summarizer in our method. The learning rate is $5e^{-5}$. We adopt learning rate warmup and decay. The vocabulary size is 50,265. We use the optimizer Adam with $\beta_1=0.9$ and $\beta_2=0.999$. The beam size of beam search used in the summary generation process is 4. We use the implementations of BART, T5, LED, and Pegasus from HuggingFace's Transformers \cite{wolf-etal-2020-transformers}. Our method and all abstractive baselines are trained on NVIDIA A100 GPU.

\subsection{Results and Discussion}
\subsubsection{Sentence Classification Results}
The prediction results of our sentence classifier on the test set of the constructed dataset \textbf{{\clsdatasetname}} are shown in Table \ref{table:sentcls_results}. We can see that the sentence classifier achieves decent F1 score for each aspect although the dataset is severely imbalanced as shown in Table \ref{table:sentclsdataset_statistics}. This indicates that the classifier helps keep the most informative sentences for each aspect and get rid of the most irrelevant ones. 
\begin{table}[ht]
\small
\centering
\resizebox{.45\textwidth}{!}{
\begin{tabular}{p{0.95cm}<{\centering} p{1.3cm}<{\centering} p{1.1cm}<{\centering} p{0.70cm}<{\centering} p{0.80cm}<{\centering}} 
 \hline
 & \textbf{Precision} & \textbf{Recall} & \textbf{F1} & \textbf{Support} \\ [0.5ex] 
 \hline
 Abstract & 0.381 & 0.280 & 0.323 & 182 \\ [0.5ex]
 Problems & 0.387 & 0.280 & 0.325 & 421 \\ [0.5ex]
 Actions & 0.378 & 0.383 & 0.380 & 81 \\ [0.5ex]
 Decisions & 0.356 & 0.415 & 0.383 & 407 \\ [0.5ex]
\hline
\end{tabular}
}
\caption{The results of sentence classification on the constructed dataset \textbf{{\clsdatasetname}}.}
\label{table:sentcls_results}
\end{table}

\begin{table*}[h]
\centering
\resizebox{1.0\textwidth}{!}{
\begin{tabular}{l| c c c| c c c| c c c| c c c}
 \hline
		\multirow{2}{*}{Models}    &\multicolumn{3}{c|}{Abstract} & \multicolumn{3}{c|}{Problem} & \multicolumn{3}{c|}{Action} & \multicolumn{3}{c}{Decision} \\ \cline{2-13} 
		& R-1 & R-2 & R-L & R-1 & R-2 & R-L & R-1 & R-2 & R-L & R-1 & R-2 & R-L \\ [0.5ex] 
 \hline
 \textbf{TextRank} \cite{mihalcea2004textrank} & 29.76 & 4.28 & 14.79 & 16.44 & 1.68 & 9.25 & 13.24 & 2.67 & 9.57 & 21.19 & 1.59 & 13.36 \\ [0.5ex]
 \textbf{LexRank} \cite{erkan2004lexrank} & 29.71 & 4.34 & 13.93 & 16.92 & 2.24 & 9.39 & 9.03 & 2.27 & 6.53 & 21.77 & 2.33 & 13.22 \\ [0.5ex]
 \hline
 \textbf{T5}\textsubscript{base} \cite{raffel2020exploring} & 27.39 & 6.25 & 22.93 & 8.66 & 1.30 & 6.33 & 22.01 & 3.76 & 17.59 & 20.59 & 6.35 & 19.03 \\ [0.5ex]
 \textbf{T5}\textsubscript{large} \cite{raffel2020exploring} & 32.05 & 8.31 & 25.73 & 5.40 & 0.20 & 3.95 & 20.66 & 5.57 & 18.04 & 21.12 & 6.29 & 19.78 \\ [0.5ex]
\textbf{Pegasus} \cite{zhang2020pegasus} & 25.3 & 4.3 & 14.5 & - & - & - & - & - & - & - & - & - \\ [0.5ex]
 \textbf{LED}\textsubscript{base} \cite{beltagy2020longformer} & 29.96 & 9.34 & 24.15 & 4.76 & 0.00 & 4.76 & 13.92 & 5.48 & 12.39 & 18.19 & 6.71 & 17.69 \\ [0.5ex]
 \textbf{BART}\textsubscript{base} \cite{lewis-etal-2020-bart} & 43.96 & 15.99 & \textbf{27.25} & 4.76 & 0.00 & 4.76 & 21.05 & 0.00 & \textbf{21.05} & 10.83 & 2.39 & 8.53 \\ [0.5ex]
 \textbf{BART}\textsubscript{large} \cite{lewis-etal-2020-bart} & \textbf{46.66} & \textbf{17.12} & 27.07 & 6.03 & 0.00 & 5.71 & 21.05 & 0.00 & \textbf{21.05} & 24.93 & 11.72 & 20.88 \\ [0.5ex]
 \textbf{ChatGPT}\textsubscript{gpt-3.5-turbo} & 25.44 & 5.79 & 22.61 & 17.20 & 2.98 & 14.81 & 10.59 & 1.28 & 9.69 & 20.40 & 3.50 & 18.72 \\ [0.5ex]
\hline
\textbf{{\methodname}} (ours) & 40.32 & 14.02 & 23.62 & \textbf{25.35} & \textbf{10.65} & \textbf{17.52} & \textbf{25.93} & \textbf{15.06} & 20.89 & \textbf{28.86} & \textbf{11.77} & \textbf{21.49} \\ [0.5ex]
\hline
\end{tabular}
}
\caption{Result comparison of aspect-based summary generation between our method and baselines.}
\label{table:exp_results}
\end{table*}

\begin{table*}[h]
\centering
\resizebox{1.0\textwidth}{!}{
\begin{tabular}{l| c c c| c c c| c c c| c c c}
 \hline
		\multirow{2}{*}{Models}    &\multicolumn{3}{c|}{Abstract} & \multicolumn{3}{c|}{Problem} & \multicolumn{3}{c|}{Action} & \multicolumn{3}{c}{Decision} \\ \cline{2-13} 
		& R-1 & R-2 & R-L & R-1 & R-2 & R-L & R-1 & R-2 & R-L & R-1 & R-2 & R-L \\ [0.5ex] 
 \hline
 Oracle(SimCSE-BERT-0.4) & 51.37 & 20.37 & 30.03 & 29.81 & 4.79 & 26.66 & 42.87 & 22.92 & 36.31 & 40.08 & 18.33 & 29.44 \\ [0.5ex]
 Oracle(SimCSE-BERT-0.46) & 48.58 & 19.55 & 30.19 & 33.67 & 5.57 & 31.38 & 49.88 & 25.40 & 40.86 & 35.50 & 17.97 & 28.03 \\ [0.5ex]
 Oracle(SimCSE-BERT-0.5) & 48.48 & 20.27 & 29.31 & 35.05 & 6.22 & 28.86 & 47.28 & 22.92 & 43.67 & 37.01 & 16.92 & 27.35 \\ [0.5ex]
 Oracle(SimCSE-RoBERTa-0.46) & 50.16 & 21.21 & 30.24 & 32.61 & 5.83 & 28.94 & 53.42 & 26.80 & 45.32 & 38.87 & 21.11 & 30.26 \\ [0.5ex]
\hline
Oracle\textsubscript{SingleModel}(SimCSE\textsubscript{RoBERTa-0.46}) & 37.92 & 13.76 & 21.89 & 39.29 & 9.66 & 32.95 & 52.62 & 20.18 & 44.97 & 36.81 & 15.50 & 26.13 \\ [0.5ex]
\hline
\textbf{{\methodname}} (ours) & 40.32 & 14.02 & 23.62 & 25.35 & 10.65 & 17.52 & 25.93 & 15.06 & 20.89 & 28.86 & 11.77 & 21.49 \\ [0.5ex]
\hline
\end{tabular}
}
\caption{Comparison between our method's performance on aspect-based summary generation and oracle results.}
\label{table:oralce_results}
\end{table*}

\begin{table}[ht]
\small
\centering
\resizebox{0.5\textwidth}{!}{
\begin{tabular}{l| c c c| c c c}
 \hline
		\multirow{2}{*}{Models}    &\multicolumn{3}{c|}{Problem} & \multicolumn{3}{c}{Decision} \\ \cline{2-7} 
		& R-1 & R-2 & R-L & R-1 & R-2 & R-L \\ [0.5ex] 
 \hline
 \textbf{BART}\textsubscript{large} & 11.47 & 1.14 & 9.52 & 19.43 & 2.11 & 13.61 \\ [0.5ex]
 \textbf{ChatGPT}\textsubscript{gpt-3.5-turbo} & 15.84 & \textbf{2.08} & \textbf{14.09} & 17.23 & 2.58 & \textbf{15.17} \\ [0.5ex]
\hline
\textbf{{\methodname}} (ours) & \textbf{16.75} & 2.02 & 10.93 & \textbf{20.75} & \textbf{3.13} & 13.33 \\ [0.5ex]
\hline
\end{tabular}
}
\caption{Performance comparison of our method, BART\textsubscript{large} and ChatGPT on the ICSI-Test set.}
\label{table:exp_results_icsi}
\end{table}

\begin{table*}[t]
\centering
\resizebox{1.0\textwidth}{!}{
\begin{tabular}{l| c c c c| c c c c| c c c c| c c c c}
 \hline
		\multirow{2}{*}{}    &\multicolumn{4}{c|}{Abstract} & \multicolumn{4}{c|}{Problem} & \multicolumn{4}{c|}{Action} & \multicolumn{4}{c}{Decision} \\ \cline{2-17} 
		& Win & Lose & Tie & Kappa & Win & Lose & Tie & Kappa & Win & Lose & Tie & Kappa & Win & Lose & Tie & Kappa \\ [0.5ex] 
 \hline
 Correctness & 60.7\% & 17.9\% & 21.4\% & 0.211 & 50.0\% & 4.8\% & 45.2\% & 0.415 & 27.4\% & 39.3\% & 33.3\% & 0.579 & 38.1\% & 20.2\% & 41.7\% & 0.417 \\ [0.5ex]
 Non-Hallucination & 48.8\% & 27.4\% & 23.8\% & 0.206 & 34.5\% & 3.6\% & 61.9\% & 0.216 & 33.3\% & 39.3\% & 27.4\% & 0.242 & 34.5\% & 20.2\% & 45.2\% & 0.175 \\ [0.5ex]
 Fluency & 20.2\% & 1.2\% & 78.6\% & 0.071 & 0.0\% & 0.0\% & 100\% & - & 10.7\% & 27.4\% & 61.9\% & 0.072 & 0.0\% & 0.0\% & 100\% & - \\ [0.5ex]
 Non-Redundancy & 3.6\% & 3.6\% & 92.9\% & 0.178 & 0.0\% & 0.0\% & 100\% & - & 10.7\% & 26.2\% & 63.1\% & 0.027 & 0.0\% & 0.0\% & 100\% & - \\ [0.5ex]
\hline
\end{tabular}
}
\caption{Human evaluation results. "Win" means the generated summary of our {\methodname} method is better than that of BART\textsubscript{large} in one perspective.}
\label{table:humaneval_results}
\end{table*}

\begin{table*}[t]
\centering
\resizebox{1.0\textwidth}{!}{
\begin{tabular}{l| c c c| c c c| c c c| c c c}
 \hline
		\multirow{2}{*}{Models}    &\multicolumn{3}{c|}{Abstract} & \multicolumn{3}{c|}{Problem} & \multicolumn{3}{c|}{Action} & \multicolumn{3}{c}{Decision} \\ \cline{2-13} 
		& R-1 & R-2 & R-L & R-1 & R-2 & R-L & R-1 & R-2 & R-L & R-1 & R-2 & R-L \\ [0.5ex] 
 \hline
 \textbf{{\methodname} (oracle)} & 33.27 & 11.12 & 20.35 & 23.43 & 9.45 & 18.00 & 50.38 & 17.86 & 43.43 & 31.83 & 11.30 & 23.60 \\ [0.5ex]
 \hline
 \textbf{{\methodname} (filtertrain-0.5)} & 40.32 & 14.02 & 23.62 & 25.35 & 10.65 & 17.52 & 25.93 & 15.06 & 20.89 & 28.86 & 11.77 & 21.49 \\ [0.5ex]
 \textbf{{\methodname} (filtertrain-0.3)} & 44.57 & 15.15 & 24.37 & 21.24 & 8.93 & 17.12 & 16.14 & 1.43 & 16.50 & 30.36 & 10.97 & 21.81 \\ [0.5ex]
 \textbf{{\methodname} (nofiltering)} & 37.55 & 12.41 & 22.17 & 19.38 & 5.06 & 14.38 & 35.88 & 19.72 & 29.66 & 31.52 & 11.69 & 23.02 \\ [0.5ex]
 \textbf{{\methodname} (down-sampling)} & 32.36 & 12.36 & 20.15 & 23.43 & 5.00 & 17.62 & 28.61 & 5.60 & 26.03 & 29.62 & 10.10 & 21.99 \\ [0.5ex]
\hline
\end{tabular}
}
\caption{Ablation study of our method on aspect-based summary generation.}
\label{table:abstudy_results}
\end{table*}

\begin{table*}[t]
\small
\centering
\begin{tabular}{p{0.85cm}<{\centering}| p{8.1cm}<{}| p{7.9cm}<{}} 
 \hline
 & \textbf{Reference Summaries} & \textbf{Generated Summaries} \\ [0.5ex] 
 \hline
 Abstract & \textbf{The project manager recapped the decisions made in the previous meeting.} \textbf{The marketing expert discussed his findings from trend watching reports, stressing the need for a product that has a fancy look and feel, is technologically innovative, is easy to use, and incorporates some aspect of fruit and vegetables in its design.} The interface specialist presented research on the appearance of current remotes, concluding that \textbf{current remotes are generally unappealing in terms of look and usability}. The interface specialist presented ideas on how to remedy the unappealing aspects of current remotes in the team's design and also how to incorporate the corporate color scheme into the design. \textbf{The industrial designer presented options for materials, components, and batteries and discussed the restrictions involved in using certain materials.} The team then discussed options for the shape and color scheme of the remote, as well as what components, batteries, and materials to use. & \textbf{The project manager opened the meeting by going over the minutes from the previous meeting.} \textbf{The industrial designer} discussed the interior workings of a remote control and \textbf{discussed options for batteries, materials, and components.} The user interface designer discussed user requirements and suggested that \textbf{the remote should have a fancy look and feel}, but not too fancy. \textbf{The marketing expert discussed trends in fashion and discussed how to incorporate them into the design of the remote control.} The interface specialist also discussed the possibility of \textcolor{red}{using a scroll wheel to help users find their remote when misplaced}, and suggested using a fruit and vegetable theme for the remote. The team then \textcolor{red}{discussed their target group and decided that the target group should be people aged fifteen to thirty-five.} The group then discussed other features to consider in making the remote, such as an LCD display, \textcolor{red}{an ergonomic design, and whether to include speech recognition.} \\ [0.5ex]
 \hline
 Problem & What sort of battery to use. \textbf{How to incorporate a fruit and vegetable theme}, the corporate color, and soft materials \textbf{into the remote.} What shape the remote should be. Whether the remote should have a talk-back function. Whether to have an LCD display on the remote. What the case will look like. Whether to include a jog dial on the remote. & \textbf{How to incorporate a fruit and vegetable theme into the design of the remote.} \\ [0.5ex]
 \hline
 Action & Some team members will design a prototype of the remote. & NA. \\ [0.5ex]
 \hline
 Decision & \textbf{The remote will be curvy.} The remote will be shaped like a snowman. The remote will not have a talk-back function. The remote will have a jog dial. \textbf{The materials used will be plastic and rubber.} The case will have a soft fruit-like feel to it. The remote will not have an LCD display. The remote will have a kinetic battery. The remote will have rubber \textbf{buttons.} & The group decided to \textcolor{red}{use a wheel for the remote.} \textcolor{red}{The remote will have an LCD screen.} \textbf{The case will be double-curved and made of rubber.} \textbf{The buttons} will be large. \\ [0.5ex]
\hline
\end{tabular}
\caption{Case study: Comparison between the generated summaries for different aspects by our method and the reference summaries.}
\label{table:case_study}
\end{table*}

\begin{table*}[ht]
\small
\centering
\begin{tabular}{p{0.95cm}<{\centering}| p{12.0cm}<{}} 
 \hline
 & \textbf{Instructions for ChatGPT} \\ [0.5ex] 
 \hline
 Abstract & Please summarize the following meeting transcript: \\ [0.5ex]
 \hline
 Problem & Please tell me the problems needed to be addressed in the following meeting transcript: \\ [0.5ex]
 \hline
 Action & Please list the actions discussed in the following meeting transcript: \\ [0.5ex]
 \hline
 Decision & Please tell me the decisions made in the following meeting: \\ [0.5ex]
\hline
\end{tabular}
\caption{Instructions for ChatGPT (gpt-3.5-turbo) to generate summaries for different aspects.}
\label{table:case_study_instructions}
\end{table*}

\begin{table*}[h]
\small
\centering
\begin{tabular}{p{0.85cm}<{\centering}| p{16.7cm}<{}} 
 \hline
 & \textbf{Reference Summaries} \\ [0.5ex] 
 \hline
 Abstract & The project manager recapped the decisions made in the previous meeting. The marketing expert discussed his findings from trend watching reports, stressing \textbf{the need for a product that has a fancy look and feel, is technologically innovative, is easy to use, and incorporates some aspect of fruit and vegetables in its design.} The interface specialist presented research on the appearance of current remotes, concluding that current remotes are generally unappealing in terms of look and usability. The interface specialist presented ideas on how to remedy the unappealing aspects of current remotes in the team's design and also how to \textbf{incorporate the corporate color scheme into the design}. The industrial designer presented options for materials, components, and batteries and discussed the restrictions involved in using certain materials. The team then discussed options for \textbf{the shape and color scheme of the remote, as well as what components, batteries, and materials to use}. \\ [0.5ex]
 \hline
 Problem & \textbf{What sort of battery to use}. \textbf{How to incorporate a fruit and vegetable theme, the corporate color, and soft materials into the remote}. What shape the remote should be. Whether the remote should have a \textbf{talk-back} function. Whether to have an LCD display on the remote. \textbf{What the case will look like}. Whether to include a jog dial on the remote. \\ [0.5ex]
 \hline
 Action & Some team members will design a prototype of the remote. \\ [0.5ex]
 \hline
 Decision & \textbf{The remote will be curvy.} The remote will be shaped like a snowman. The remote will not have a talk-back function. The remote will have a jog dial. \textbf{The materials used will be plastic and rubber.} The case will have a soft fruit-like feel to it. The remote will not have an LCD display. The remote will have a kinetic battery. The remote will have \textbf{rubber buttons.} \\ [0.5ex]
\hline
\end{tabular}
\caption{Case study for ChatGPT generated summaries: The part of reference summaries.}
\label{table:case_study_groundtruth}
\end{table*}

\begin{table*}[t]
\small
\centering
\begin{tabular}{p{0.95cm}<{\centering}| p{14.6cm}<{}} 
 \hline
 & \textbf{ChatGPT's Outputs} \\ [0.5ex] 
 \hline
 Abstract & \textcolor{red}{The speaker shares the progress of the project and reminds everyone that the meeting's goal is to make a decision on the product concept. They discuss their target group, decided that the product will be non-rechargeable} \textbf{battery}-powered, and talk about \textbf{incorporating corporate color and slogan}. The speaker shares their background research on the remote control market, \textbf{highlighting that a fancy look and feel is a top priority. They suggest thinking about the shape, materials, and themes or series} that unify the product. \textcolor{red}{The speaker shares some examples and suggests thinking about how to incorporate the LCD display, speaker, and slogan. Finally, the speaker suggests thinking about a high-tech user-friendly theme to achieve project goals.} \\ [0.5ex]
 \hline
 Problem & Based on the given transcript, here are the problems that need to be addressed in the meeting: \textcolor{red}{Decision-making: The participants need to decide how to fulfill the requirements and objectives. Speech recognition: There is uncertainty regarding the decision to incorporate speech recognition into the product. Innovation and ease of use: The team needs to find a way to be innovative while ensuring the product remains user-friendly.} Incorporating fashion trends: There is a discussion about \textbf{incorporating fruit and vegetable themes into the design, but the team needs to determine how to integrate it} without compromising the overall concept. Material trends: \textbf{The team needs to consider using soft, spongy, and ergonomic materials} that are user-friendly. Design aesthetics: The existing remote controls in the market are unattractive and easily lost. The team needs to address the design to make it visually appealing, comfortable to hold, and easily identifiable. Color scheme: \textbf{The corporate color} scheme needs to be determined and its flexibility explored to ensure it aligns with the desired design. Additional features: The team needs to decide on hidden or visible additional features, including exploring the possibility of \textbf{voice activation} and alternative buttons. \textcolor{red}{Slogan and branding: The positioning statement and slogan for the product need to be determined. Component selection: The team needs to consider the cost, complexity, size, and power consumption of different components to make informed decisions.} Case design and materials: \textbf{The options for the case design}, including flat or curved, and the selection of materials need to be discussed. \textcolor{red}{Logic chips: The complexity and functionality of logic chips need to be evaluated based on the desired functions of the remote control.} Power options: The team needs to decide on the power source, \textbf{considering alternatives such as batteries, solar cells, and kinetic energy}. \textcolor{red}{User interface restrictions: Certain features may require more complex logic or have limitations based on the chosen materials, such as the inability to have solar panels with a rubber case. Production considerations: Simplifying components and reducing costs while maintaining quality and profitability should be a priority.} These problems and discussions provide a starting point for the meeting agenda to address various aspects of the remote control design and functionality. \\ [0.5ex]
 \hline
 Action & \textcolor{red}{Actions discussed in the meeting: Recap on the minutes from the last meeting. Reach a decision on the concepts of the product. Summarize market research and identify trends. Discuss design options to inform other elements of the project. Focus on innovative design technology that is cutting-edge yet easy to use. Think about shape, materials, and themes that unify the marketing identity. Confirm any action and possibility with the team. Work through the components that will have an impact on the project. Bear in mind the ultimate goal of the meeting to reach a decision on the concepts of the product. Think about a concept that touches back to the company's color (yellow). Consider the support for the product. Plan for the next meeting in thirty minutes.} \\ [0.5ex]
 \hline
 Decision & Based on the given conversation, here are the decisions made in the meeting: \textcolor{red}{Ease of use and innovative design technology are essential. Incorporate vibrant natural colors}, \textbf{softness in materials}, \textcolor{red}{shape, and function. Consider incorporating features from a Mac iPod, but with a more comfortable and user-friendly design. Explore the idea of incorporating lighting within the remote control, such as buttons lighting up in different colors. Focus on large, easy-to-press buttons that are comfortable to use. Hide any extra features beyond the basic ones, making them visible only when opened or with special effort. Avoid the traditional "big square block" look of current remote controls}. Consider a \textbf{curved, ergonomic shape for the remote control}. \textcolor{red}{Yellow color should be incorporated into the design}. \textbf{Use materials like plastic and rubber} or \textcolor{red}{wood and titanium, but avoid mixing them}. \textcolor{blue}{Explore different power source options: standard battery, kinetic (movement-powered), solar, or wind-up}. \textcolor{red}{Consider a strong battery to power a light in the remote control. Discuss the position of the slogan on the remote control.} \textcolor{blue}{Determine the need for an LCD display and speaker on the remote control}. \textcolor{red}{Position the buttons on the front of the remote control. Consider a transparent case with strategically placed lights for a subtle glow effect.} \textbf{Use rubber buttons} for a comfortable feel. \textcolor{blue}{Explore the possibility of incorporating a jog dial for certain functions}. \textcolor{red}{Ensure the remote control is user-friendly for channel selection and skipping. Aim for a quick and efficient user experience. Please note that some parts of the conversation were ambiguous, and the decisions listed are based on the available information}. \\ [0.5ex]
\hline
\end{tabular}
\caption{Case study for ChatGPT generated summaries: The part of ChatGPT's outputs.}
\label{table:case_study_chatgpt}
\end{table*}

\subsubsection{Aspect-based Summarization Results}
\noindent{\bf $\bullet$ \emph{Comparison with Baselines.}}
The experimental results of our method and baselines on aspect-based summarization are shown in Table \ref{table:exp_results}. We train the same baseline model for each of the aspects and use the results obtained from the separate models for each aspect. For example, for the baseline BART\textsubscript{large}, we use the same model architecture to train four different models, each of which is trained to produce the corresponding aspect-based summary. In contrast, our method only has one single summarizer which is trained for producing summaries for all aspects. From Table \ref{table:exp_results}, one can see that our method performs much better than all the baselines on the three aspects including Problem, Action and Decision. Although its performance on abstract summary is not as good as the strong baselines such as BART, it is still reasonably better than other baselines. The reason for the performance drop of our method on the abstract summary is that there is some information loss in the long meeting transcript. This loss of information is valuable for creating a more comprehensive abstract, as abstracts require a broader range of information from the entire meeting to be considered complete.
The better performance on the other three aspects (i.e., problem, action and decision) verifies that our two-stage method can help select the most informative sentences for different aspects in the meeting. In other words, the sentence classifier helps improve aspect-based meeting transcript summarization.

\noindent{\bf $\bullet$ \emph{Comparison with Oracle Results.}}
In the first stage of our method, we utilized the reference summaries for all aspects of the meeting transcript to construct a sentence classification dataset for training our classifier, which will be used in the second stage to help improve the performance of aspect-based summary generation. To investigate the effect of the sentence classifier and how much it helps, we use the reference summaries of all aspects in the test set to filter our irrelevant sentences in the input meeting transcript and then feed the filtered transcript in the second stage to see the upper bound of the performance of aspect-based summary generation. In this way, we can see the performance gap between our method and the oracle results. For the oracle results calculation, in both the training and test set, we obtain the embedding of each sentence in the meeting transcript and the embedding of reference summary for different aspects. Then for each aspect, we calculate the semantic similarity between each sentence and the aspect summary by using their embeddings. Each sentence with a similarity score higher than the threshold $\alpha$ will be kept in the meeting transcript, otherwise it will be discarded. Those filtered meeting transcripts for different aspects are used to train four different summarizers to produce a summary for each aspect. The different sentence embedding models and different values of the threshold $\alpha$ will impact the oracle results. Therefore we also show the results of different combinations of embedding model and value of $\alpha$ in the comparison as shown in Table \ref{table:oralce_results}. The first four rows of Table \ref{table:oralce_results} show the oracle results of four separate summarizers
for producing the summary of each aspect. The fifth row shows the oracle results of one single summarizer
producing summaries for all aspects. The performance gap between the fifth and last row shows that there is still a large space to improve the performance of the sentence classifier in our method, although it helps improve aspect-based meeting transcript summarization as discussed before.

\noindent{\bf $\bullet$ \emph{More Results on the ICSI-Test Set.}}
To further evaluate our method’s effectiveness and generalizability, we tested our method and two strong baselines on another test set derived from the ICSI corpus (containing 61 testing examples). The results are shown in Table \ref{table:exp_results_icsi}. Our method generally outperforms the baselines on R-1 and R-2. ChatGPT achieves higher R-L scores, which may be attributed to the generation of lengthy aspect-based summaries as shown in Table \ref{table:case_study_chatgpt}.

\noindent{\bf $\bullet$ \emph{Human Evaluation.}}
We also conducted a human evaluation to compare aspect-based summaries generated by different models in terms of correctness (factual consistency with ground-truth summaries), non-hallucination (less fake or incorrect information), fluency (content organization and grammar), and non-redundancy (minimization of repetitive information). We used 21 examples from the AMI test set. Four annotators are asked to compare the aspect-based summaries produced by two models, which are presented anonymously. 
Fleiss' kappa \cite{fleiss1971measuring} is utilized to assess their agreements. 
Table \ref{table:humaneval_results} shows that our method
outperforms the strong baseline BART\textsubscript{large} in terms of correctness and non-hallucination for aspects such as Abstract, Problem and Decision. 
{\methodname} achieves comparable results to BART\textsubscript{large} on summarization for Action. Both methods exhibit similar fluency and non-redundancy results across all aspects, which is reasonable since they are both based on BART.

\noindent{\bf $\bullet$ \emph{Ablation Study.}}
Most of the sentence examples in the constructed dataset {\clsdatasetname} are irrelevant to any of the aspects in the meeting. To train the sentence classifier much better and quicker, we developed different filtering strategies for our method, which are presented as follows.
1) \textbf{{\methodname} (filtertrain-0.5)}. It only removes sentences without any aspect label in the training set of {\clsdatasetname}. The threshold used to predict the aspect label for each sentence in the test set is 0.5. 2)\textbf{{\methodname} (filtertrain-0.3)}. It is similar to {\methodname} (filtertrain-0.5), the only difference is that the threshold for aspect label prediction is 0.3.
3)\textbf{{\methodname} (nofiltering)}. It does not remove any sentence in {\clsdatasetname}.
4)\textbf{{\methodname} (down-sampling)}. We make a balanced training set in {\clsdatasetname} to train the sentence classifier by conducting down sampling on the training set to reduce the number of irrelevant sentence examples. The down-sampled training set only contains 3367 irrelevant sentence examples, which makes it more balanced. 
5)\textbf{{\methodname} (oracle)}. It removes sentences without any aspect label in both training and test set of {\clsdatasetname}. It represents the upper bound performance of our method's filtering strategy for aspect-based summary generation. The results of different filtering strategies are shown in Table \ref{table:abstudy_results}.
We can see that filtering sentences in the training set helps improve the summary generation performance for some aspects significantly, including "Abstract" and "Problem", but causes a performance drop in the "Action" aspect. And it does not affect the performance of summary generation for the aspect "Decision". This shows that different aspects have a different relationship with the sentences without any aspect label. The sentence classifier still does not select the most important sentences for the "Action" aspect. There is still much space to improve the performance of the sentence classifier.
The results of down-sampling on "Abstract" and "Problem" are comparable to the oracle results (the first row in Table \ref{table:abstudy_results}), which indicates that down-sampling helps improve the summary generation performance for those aspects. However, it does not obviously benefit the generation performance in other aspects. In summary, \textbf{{\methodname} (filtertrain-0.5)} and \textbf{{\methodname} (nofiltering)}  
generally yield better results than other approaches. 
We use the results of \textbf{{\methodname} (filtertrain-0.5)} 
to compare our method with baselines.

\subsubsection{Case Study}
We conduct a case study to show whether our method can generate good summaries for different aspects of the meeting transcript. 
Table \ref{table:case_study} shows the generated and reference summaries for a meeting transcript in the test set. The bold parts in the generated summaries mean that they exist in the reference summaries. In other words, they are generated correctly for the corresponding aspects. Those parts in red are the content not mentioned (e.g., using a scroll wheel to help users find their remote when misplaced) in the meeting or the content mentioned but summarized incorrectly (e.g., the remote will have an LCD screen) in the generated aspect summaries. This indicates that our method can help improve the performance of aspect-based summary generation by a certain degree. However, there is still incorrect or fake content generated in the aspect summaries. We will explore more techniques in the future to produce more accurate 
aspect-based summaries.  
\subsubsection{Case Study for ChatGPT Generated Summaries}
\label{sec:appendix_chatgpt}
Besides evaluating the effectiveness of ChatGPT (gpt-3.5-turbo) in generating aspect-based summaries for meeting transcripts in Table \ref{table:exp_results} and \ref{table:exp_results_icsi}, 
we also conduct a case study to show the quality of its generated aspect-based summaries. 
We use different instructions combined with the meeting transcript as the input for ChatGPT as shown in Table \ref{table:case_study_instructions}. 
The results and summaries of ChatGPT are obtained by using OpenAI API service during May 10-22, 2023.
Table \ref{table:case_study_chatgpt} exhibits the summaries generated by ChatGPT for different aspects. Table \ref{table:case_study_groundtruth} shows the corresponding ground-truth summaries for different aspects for the selected meeting transcript. The bold parts in both Tables represents the overlap of content between the ground-truth summaries and ChatGPT's outputs. The red parts in Table \ref{table:case_study_chatgpt} are incorrect content which are not mentioned in the ground-truth summaries or mentioned incorrectly. Interestingly, the blue parts in the ChatGPT generated decision-based summary do not make a decision on the different choices of some feature of the remote control. They are actually more like the problems discussed in the meeting. From these two tables, one can see that ChatGPT has limited ability of generating correct summaries for different aspects. Because a meeting transcript is lengthy and complex, and the sentences relevant to different aspects mingle together and scatter in the long meeting transcript. This makes it challenging to generate correct aspect-based summaries for meeting transcripts. There is still much room to improve the aspect-based meeting transcript summarization by designing more effective methods.

\section{Conclusion}
In this paper, we propose a new task of aspect-based meeting transcript summarization, which aims to generate the summary individually for each aspect of the meeting content. To identify the salient information from mixed and long meeting content, we propose a two-stage method, which first selects the meeting transcript sentences related to each aspect, then merges the selected sentences as the input of the summarizer to produce the aspect-based summary. Experiments on the AMI corpus show that our method outperforms competitive baselines, which verifies its effectiveness.

\section*{Acknowledgment}
We thank the reviewers for their feedback. This work is supported in part by NSF under grant III-2106758.

\vspace{12pt}

\end{document}